\documentclass[9pt,twocolumn,twoside]{opticajnl}

\journal{ol} 

\setboolean{shortarticle}{true}

\usepackage{lineno}
\usepackage{soul}

\renewcommand{\hl}{}
\renewcommand{\ul}{}
\nolinenumbers

\title{Addressing Data Scarcity in Optical Matrix Multiplier Modeling Using Transfer Learning}

\author[1]{Ali Cem}
\author[1]{Ognjen Jovanovic}
\author[2]{Siqi Yan}
\author[1]{Yunhong Ding}
\author[1]{Darko Zibar}
\author[1,*]{Francesco Da Ros}

\affil[1]{DTU Electro, Technical University of Denmark, Kongens Lyngby, Denmark, 2800}
\affil[2]{School of Optical \& Electronic Information, Huazhong Univ. of Science and Technology, Wuhan, China, 430074}

\affil[*]{Corresponding author: fdro@dtu.dk}

\begin{abstract}
We present and experimentally evaluate using transfer learning to address experimental data scarcity when training neural network (NN) models for Mach-Zehnder interferometer mesh-based optical matrix multipliers. Our approach involves pre-training the model using synthetic data generated from a less accurate analytical model and fine-tuning it with experimental data. Our investigation demonstrates that this method yields significant reductions in modeling errors compared to using an analytical model, or a standalone NN model when training data is limited. Utilizing regularization techniques and ensemble averaging, we achieve $<1$ dB root-mean-square error on the \ul{3$\times$3} matrix weights implemented by a photonic chip while using only $25\%$ of the available data.
\end{abstract}

\setboolean{displaycopyright}{true}

\begin{document}
\maketitle


Significant transformations have taken place in the field of machine learning (ML) recently due to advancements in computational hardware and the achievements of deep learning. However, with the approach of the post-Moore era, conventional computers are unlikely to meet the growing computational demands \cite{shastri2021photonics}. Neuromorphic engineering aims to develop new hardware architectures that match the distributed nature of machine learning algorithms, leading to faster and more energy-efficient computation. To overcome limitations in traditional electronic systems, photonic integrated circuits (PICs) are being explored for implementing neural networks (NNs). PICs have higher bandwidths and lower losses compared to electrical circuits, making them well-suited for NN implementations, particularly for the massively-connected linear layers \cite{zhou2022photonic}. Various optical architectures, including Mach-Zehnder interferometer (MZI) meshes \cite{shen2017deep, de2021photonic}, microring weight banks \cite{zhang2022silicon}, and photonic crossbar arrays \cite{youngblood2022coherent} have been proposed to harness the potential of PICs for efficient computation.


MZI meshes are commonly employed for optical matrix multiplication (OMM) to implement linear layers in feedforward optical NNs. \hl{The implemented linear weights are typically adjusted using voltages applied to thermo-optic phase shifters}~\cite{shen2017deep}.  \hl{Two directions are conventionally followed to accurately program an MZI mesh: online (\textit{in-situ}) and offline (\textit{in-silico}) training}~\cite{cem2023data}\hl{. Whereas several in-situ methods have been proposed}~\cite{pai2023experimentally,bandyopadhyay2021hardware} \hl{and a few experimentally validated}~\cite{pai2023experimentally}
\hl{, a calibration stage or initialization through an offline model is normally considered}~\cite{bandyopadhyay2021hardware}. It is therefore important to have a model that relates the matrix weights to the heater voltages. While simple physics-based models based on analytical expressions of MZI transmission are commonly used, they may suffer from high modeling errors due to fabrication imperfections and difficult-to-model effects like deterministic thermal crosstalk between heaters \cite{fang2019design}. \hl{These effects could potentially be included but would yield extremely evolved physical models, challenging to train.} Despite the fact that various techniques have been developed to improve the accuracy of MZI mesh programming~\cite{bandyopadhyay2021hardware}, thermal crosstalk remains a limiting factor for scaling up the size of OMMs, as its impact increases with the number of MZIs per PIC footprint \cite{milanizadeh2020control}. Recently, machine learning methods have demonstrated the ability to model MZI meshes accurately, even in the presence of fabrication tolerances and crosstalk. By leveraging measurement data, NN models can predict the weights implemented by a fabricated PIC. Nevertheless, a considerably larger number of measurements is required in order for this method to surpass the performance of analytical models \cite{cem2023data}.

It is often challenging to acquire a large number of experimental measurements that can be used to train the NN models. In the field of ML, one way to address the problem of insufficient data is to apply transfer learning (TL) \cite{tan2018survey}. TL tries to transfer the knowledge from a source domain to a target domain. By training the model on the source domain, the required training data in the target domain is significantly reduced. To further improve the performance obtained using a single model, ensemble methods are commonly used in the field of ML, where the individual predictions of multiple models are combined \cite{sharkey1996combining, dong2020survey}. While this approach is more computationally demanding than training a single model, achieving the same performance on one non-ensemble model is typically an even more challenging alternative. As a result, ensemble learning has demonstrated impressive results, especially for tasks with limited data availability such as those used for ML competitions \cite{dong2020survey}.

\begin{figure*}[!t]
\centering
\includegraphics[width=0.75\linewidth]{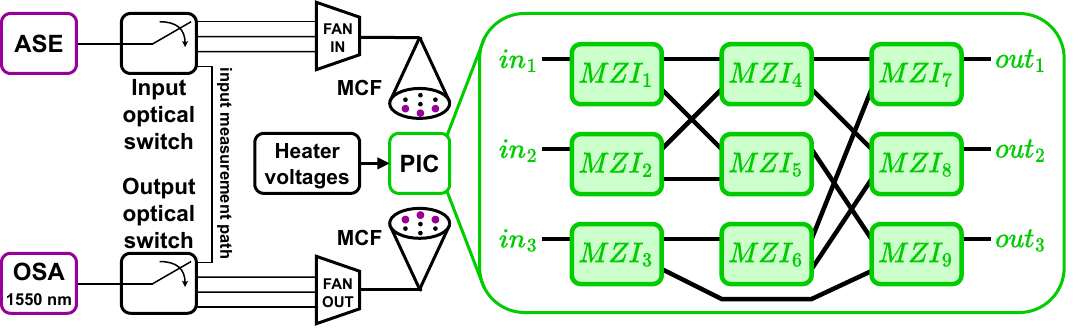}
\caption{Experimental measurement setup for OMM using the Silicon PIC alongside the MZI mesh topology \hl{(minimum MZI spacing of 100~$\mu m$)}. ASE: amplified spontaneous emission, MCF: multi-core fiber, OSA: optical spectrum analyzer.}
\label{fig_exp}
\end{figure*}

In this work, we extend upon our previous work on TL-assisted data-driven modeling of MZI meshes in \cite{cem2022data} by analyzing the performance of a simple analytical model, a standalone NN model, and a TL-assisted NN model for different numbers of experimental measurements available for training. Furthermore, we also demonstrate how a well-regularized ensemble of NN models can be used to improve the modeling errors for offline training further, enabling $<1$ dB root-mean-square error (RMSE) for a fabricated PIC while only using $25\%$ of the available data.

For the case of a MZI mesh with thermo-optic phase shifters implementing OMM, the forward model relates the variable heater voltages $\mathbf{V}$ to the implemented matrix weights $\mathbf{W}$.  When training the model offline, the optimal parameters of the model are found using experimental measurements, allowing the prediction of $\mathbf{W}$ based on $\mathbf{V}$. A simple physics-based analytical model (AM) such as the one in (\ref{eqn_model1}) is a good starting point with few trainable parameters compared to a data-driven model \cite{cem2023data}:

\begin{equation}
W_{i,j} = \alpha_{i,j} \prod_{m \in M_{i,j}} \frac{1}{4} \left| \frac{\sqrt{ER} - 1}{\sqrt{ER} + 1} \pm e^{i (\phi^{(0)}_{m} + \sum_{n=1}^{N_{MZI}} \phi^{(2)}_{m,n} V_n^2)} \right|^2 .
\label{eqn_model1}
\end{equation}

Note that $\alpha_{i,j}$ is the optical loss, $M_{i,j}$ is the set of MZIs on the path from input $j$ to output $i$, $ER$ is the MZI extinction ratio, $N_{MZI}$ is the total number of MZIs, and $\phi^{(0)}_{m}$ \& $\phi^{(2)}_{m,n}$ are the phase parameters. \hl{Optical loss, ER, and phases are optimized during training, thus partially including fabrication errors in the model}. While AM  attempts to account for thermal crosstalk through the use of the phase parameter $\phi^{(2)}_{m,n} \; (m\neq n)$, this approach falls short of reaching the modeling performance obtained using a NN model for a fabricated chip with significant thermal crosstalk \cite{cem2023data}. On the other hand, the training process of the NN model demands a substantial quantity of experimental measurements, and it has been observed that when only a limited number of measurements are available, AM surpasses the performance of the NN model \cite{cem2023data}. To address this limitation, we propose a hybrid modeling approach for scenarios with few available measurements: the TL-assisted NN model (TL-NN).

TL relaxes the assumption that training and testing data must be \hl{independent and identically distributed (i.i.d.)} and it relies on transferring knowledge from a source domain to a target domain \cite{tan2018survey}. The network-based TL approach is to reuse parts of a network pre-trained in the source domain, including network structure and weights, and fine-tune the network in the target domain. In our case, the source domain uses the synthetic dataset generated using AM and the target domain uses the experimentally measured dataset. TL-NN is trained as follows: (i) AM is trained using the available experimental data, (ii) synthetic data is generated numerically utilizing AM, (iii) the NN model is pre-trained using synthetic data, and (iv) the NN model is subsequently re-trained using the same available experimental data to enhance accuracy. Some weights and biases in the initial layer(s) of the NN are kept constant after pre-training to retain the knowledge acquired from AM.

The experimental measurement setup for OMM by a $3\times 3$ matrix is shown in Fig. \ref{fig_exp}. A broadband amplified spontaneous emission (ASE) source is used to probe the \hl{Silicon PIC}. A $1\times 4$ optical switch was employed to sequentially probe the 3 inputs of the chip, which are accessible through grating couplers. A separate input measurement path was used to record the input. The arrangement of grating couplers on the PIC necessitates probing it with a 7-core multi-core fiber (MCF). Thus, the outputs of the optical switch were connected to the 3 inputs of a fan-in device coupling from single-mode inputs to an MCF output \cite{ding2016reconfigurable}. Only 3 out of the 7 cores were used in this work. To ensure the chip operated within the linear regime, the optical power at the output of the input MCF was maintained at approximately 12 dBm. The heaters on the PIC were controlled using a printed circuit board and digital-to-analog converters. \hl{In the PIC, only one phase shifter was implemented per MZI, allowing only amplitude control and thus the implementation of a positive-only matrix.} At the output of the chip, a second set of grating couplers was used to out-couple the light onto a second MCF followed by a fan-out device, and the 3 relevant outputs were connected to a $4\times 1$ optical switch. This optical switch was further linked to an optical spectrum analyzer (OSA), where a 50 GHz spectral bandwidth at 1550 nm is selected for the measurement of the matrix weights. The measurements were carried out by fixing a set of voltages applied to the 9 MZIs under test and sequentially measuring the input optical power $P_{in}$ and the output optical power $P_{out}$ to calculate the implemented matrix weight using $W = \frac{P_{out}}{P_{in}}$ \hl{and convert them in dB scale for further use}. Additional details on the measurement procedure and the PIC itself can be found in \cite{cem2023data} and \cite{ding2016reconfigurable}, respectively.

When generating a dataset for training the models, it is necessary to measure matrix weights corresponding to various voltages. The applied voltages were initially swept from $0$ to $2$ V, covering one half-period of the MZIs' transfer function. For each MZI voltage, a sweeping process was conducted individually throughout the entire range with a step size of $0.1$ V while the remaining heater voltages were held constant, resulting in a total of $189$ measurement points. Additional measurements were conducted to increase the size of the dataset where the values for the $9$ applied voltages were sampled from $9$ i.i.d. uniform distributions from $0$ to $2$ V. In total, a training dataset $\mathcal{D}_{exp,train}$ with $4400$ sets of $\{\mathbf{V}, \mathbf{W}\}$ were used for training the models and $700$ measurements were reserved for the testing dataset $\mathcal{D}_{exp,test}$. When analyzing performance under limited data availability, fewer training points were used to train AM and two NN models with and without TL. Note that the initial $189$ measurements were always kept in $\mathcal{D}_{exp,train}$. AM was trained in MATLAB using the unconstrained multivariable optimization function \textit{fminunc()} and was used to generate $50,000$ new synthetic measurements using random voltages (uniformly distributed from $0$ to $2$ V) as inputs, resulting in the synthetic dataset $\mathcal{D}_{synth}$.

\begin{figure}[t]
\centering
\includegraphics[width=0.85\linewidth]{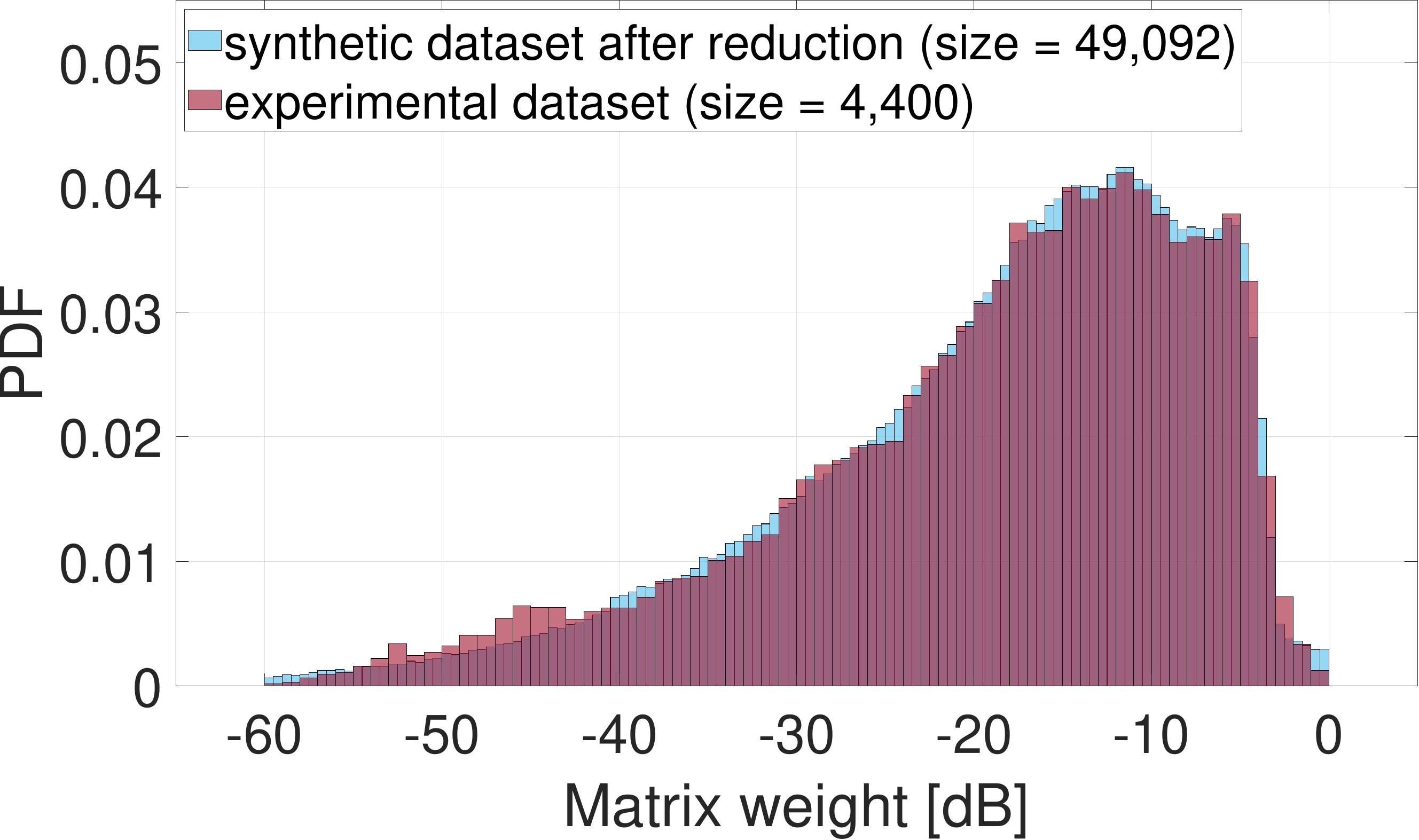}
\caption{Histograms of matrix weights $\mathbf{W}$ for the synthetic and experimental datasets for training, normalized individually to represent the estimated probability density functions (PDFs).}
\label{fig_histogram}
\end{figure}

\begin{figure}[ht]
\centering
\includegraphics[width=.75\linewidth]{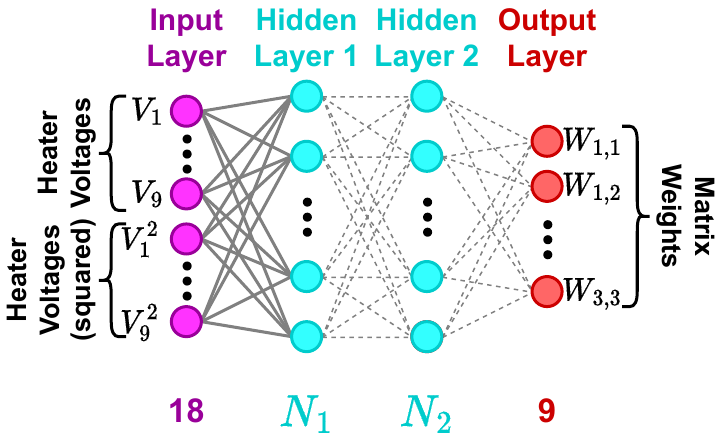}
\caption{Architecture for the NN-based PIC models, number of nodes in each layer shown below. Solid lines indicate the weights that are fixed during the re-training stage for TL-NN.}
\label{fig_nn}
\end{figure}

The histograms for the matrix weights in $\mathcal{D}_{synth}$ and $\mathcal{D}_{exp}$ are shown in Fig. \ref{fig_histogram}. $\mathcal{D}_{synth}$ initially contained matrix weights below $-60$ dB, which are not present in $\mathcal{D}_{exp}$. To ensure consistency, these datapoints were excluded from $\mathcal{D}_{synth}$ as shown in Fig. \ref{fig_histogram}, resulting in a size reduction of $<2\%$. The distributions of the two datasets exhibit remarkable similarity. The reduced $\mathcal{D}_{synth}$ was used for pre-training the NN model before re-training with experimental data. The architecture for NN and TL-NN is shown in Fig. \ref{fig_nn}. All NNs employed a hyperbolic tangent activation function, and for regularization, $L_1$ and $L_2$ regularizations were used. \hl{Both the voltages and their squares were used as input features to provide the NN with physically-related information}~\cite{cem2023data}. The objective was to minimize the RMSE between the predicted matrix weights $\mathbf{\widehat{W}}^{(l)}$ and measured matrix weights $\mathbf{W}^{(l)}$ in dB for all models, resulting in the following cost function:

\begin{flalign}
\label{eqn_rmse}
\mathcal{C}(\boldsymbol{\beta}) = &\sqrt{\frac{\sum_{i=1}^{3}\sum_{j=1}^{3}\sum_{l=1}^{L}\left(\widehat{W}^{(l)}_{i,j}(\boldsymbol{\beta})-W^{(l)}_{i,j}\right)^2}{9 \cdot L}} \\\nonumber
&+ \sum_{p=1}^{P} \lambda_{L1} \left | \beta_p \right | + \sum_{p=1}^{P} \lambda_{L2}\left ( \beta_p \right )^2 
\end{flalign}

Note that $l$ is the datapoint index, $L$ is the dataset size, $\beta_p$ is the trainable NN parameter (weight or bias) with index $p$, and $P$ is the total number of trainable parameters. The pre-training process was conducted using PyTorch with the L-BFGS optimizer. To optimize the performance of NN and TL-NN, hyperparameters including the number of nodes in the hidden layers $N_1$ and $N_2$, and the regularization parameters $\lambda_{L1}$ and $\lambda_{L2}$ were optimized using a grid search on a validation set, whose size was set to be $100$ datapoints. The values found after optimization are reported in Table \ref{tab_params}.

\begin{table}[htbp]
\centering
\caption{\bf Neural Network Hyperparameters}
\begin{tabular}{|c||c|c|c|c|}
\hline
Model (Training Data) & $N_1$ & $N_2$ & $\lambda_{L1}$ & $\lambda_{L2}$ \\
\hline
NN (400\&1000) & $83$ & $131$ & $1\times 10^{-2}$ & $1\times  10^{-4}$\\
NN (4400) & $83$ & $131$ & $2\times 10^{-4}$ & $1\times 10^{-7}$\\
TL-NN (400\&1000) & $400$ & $400$ & $5\times 10^{-4}$ & $9\times 10^{-9}$\\
\hline
\end{tabular}
  \label{tab_params}
\end{table}

\begin{figure}[!b]
\centering
\includegraphics[width=0.85\linewidth]{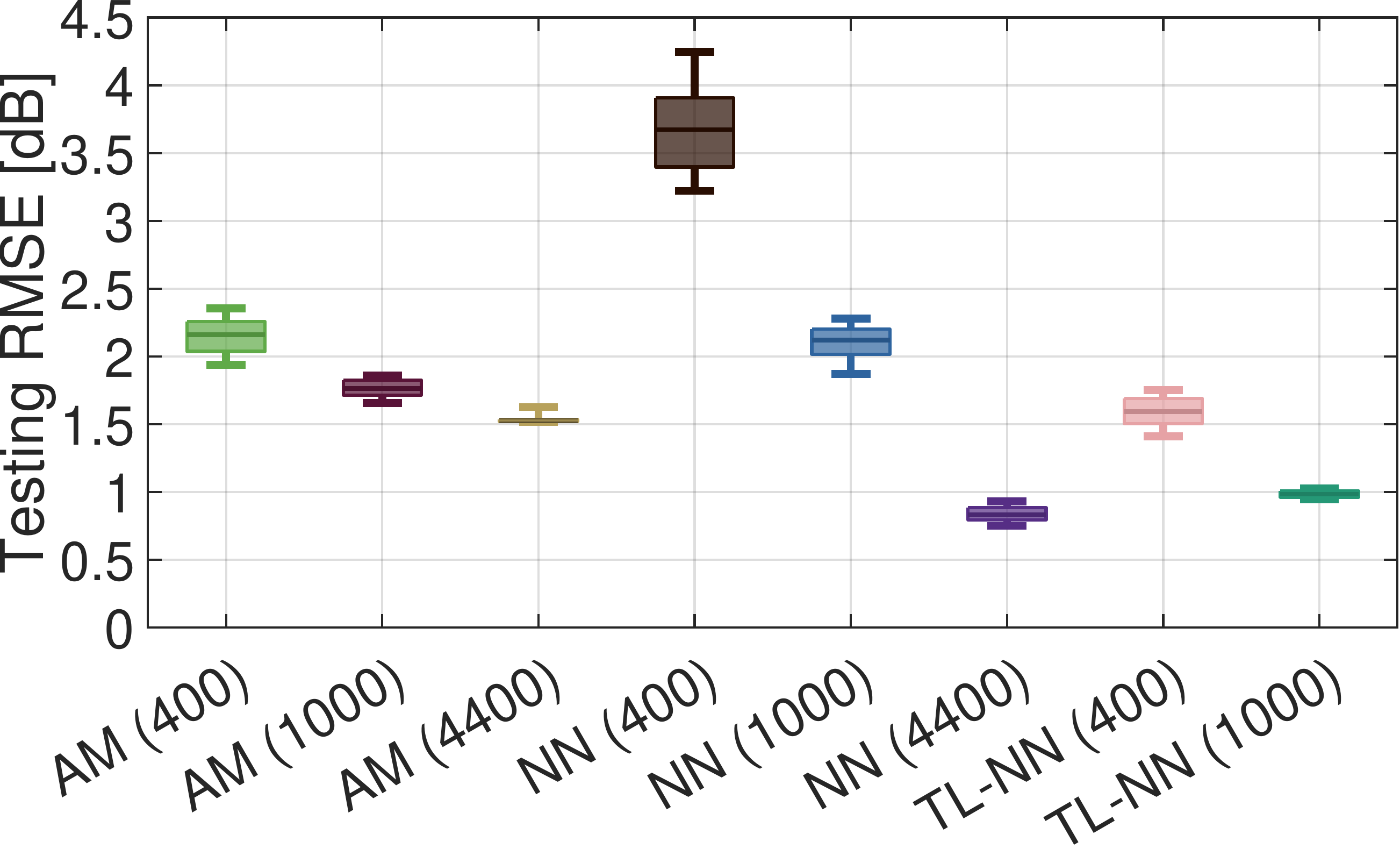}
\caption{Testing RMSEs for the models, number of experimental measurements used for training shown in parentheses. Boxes show $25^{th}$ and $75^{th}$ percentiles while the whiskers show $10^{th}$ and $90^{th}$ percentiles for $20$ random seeds.}
\label{fig_boxplot}
\end{figure}

Fig. \ref{fig_boxplot} shows the testing RMSEs for the individual models without the use of ensemble averaging. $20$ different seeds were used to initialize the NN weights and the Hessian matrix for the L-BFGS optimizer. Specifically for TL-NN, the random seed also affects the voltages used to generate $\mathcal{D}_{synth}$. In addition, the measurements were sampled randomly from all $4400$ available datapoints in $\mathcal{D}_{exp,train}$ in a fixed way for all cases where $400$ or $1000$ measurements were used for training. Comparing the results for AM and NN, the former is the better choice when training data is scarce, but its performance does not improve as much when more datapoints are available due to its limited modeling capabilities \cite{cem2022data}. A well-regularized NN model outperforms AM when more than $1000$ measurements are used for training, eventually achieving $<1$ dB RMSE when the entire experimental dataset is used. Switching over to TL-NN, the median RMSEs are strictly better than those obtained by AM for all cases. Furthermore, the RMSE is around $1$ dB even when only $1000$ measurements are used, which corresponds to approaching the performance of the NN model with a $75\%$ reduction in experimental data needed for training. \hl{Remark that an RMSE $\leq1$~dB is expected to yield negligible performance penalty when the matrix is used for inference, as shown in}~\cite{cem2022data}.

\begin{figure*}[!t]
\centering
\includegraphics[width=.9\linewidth]{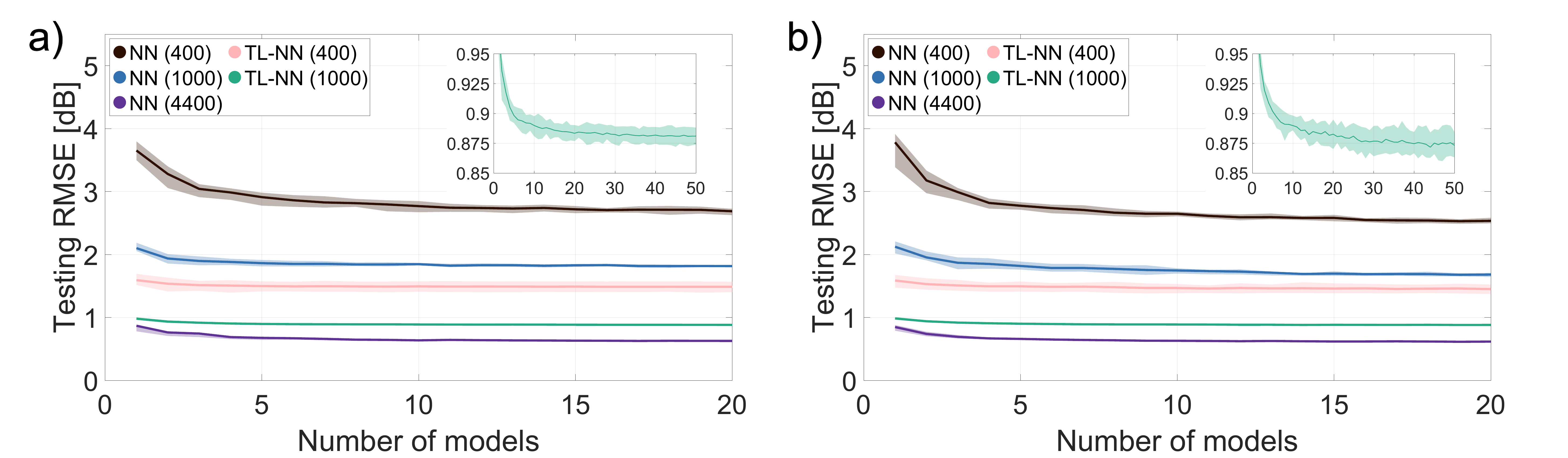}
\caption{Testing RMSEs for the ensemble NN models using a) simple averaging and b) weighted averaging. Shaded regions show $25^{th}$ and $75^{th}$ percentiles. Inset: TL-NN trained with $1000$ experimental measurements for up to $50$ models in the ensemble.}
\label{fig_ensemble}
\end{figure*}

In order to further improve the modeling performances of the individual NN-based models, ensemble averaging can be used by taking advantage of the slight variations on modeling accuracy due to the choice of the random seed. Up to $20$ different models with the same network architecture were initialized using different random seeds and they were trained on the same portion of $\mathcal{D}_{exp,train}$. Then, two linear methods were used to combine their predictions into a single final prediction: (i) simple averaging and (ii) weighted averaging \cite{sharkey1996combining}. For (ii), the weights were trained using ridge regression where the ridge parameter was optimized on the same validation set used for hyperparameter optimization. \hl{More advanced weighting techniques could potentially improve the results but only a minor impact is expected.} The results for $50$ runs per number of models in the ensemble with randomly sampled training datasets and randomly chosen random seeds for NN initialization are shown in Figure \ref{fig_ensemble}. Note that all models within the same run were trained using the same portion of $\mathcal{D}_{exp,train}$. The shaded regions show the $25^{th}$ and $75^{th}$ percentiles while the dark lines indicate the medians. The testing RMSEs improve with the number of models in the ensemble. Using $10$ models is sufficient to achieve the majority of the error reduction due to ensemble averaging, both for the simple and weighted approaches. Comparing the two, the weighted average consistently produces better results in terms of the median testing RMSE, but the improvement is within error bounds for NN (4400) and TL-NN. The number of models in the ensemble was extended to $50$ for TL-NN (1000) to better showcase the differences between simple and weighted averaging. Focusing on the insets in Fig. \ref{fig_ensemble}, we observe that weighted averaging results in lower median RMSE at the cost of higher variance, hinting towards overfitting on the validation set for the weights as more models are used in the ensemble.

In this study, we conduct an experimental evaluation of TL to address the data scarcity problem for NN-based optical matrix multiplier models. Our approach involves pre-training the NN model using synthetic data generated by an analytical model with lower accuracy. The results of our investigation demonstrate that our proposed method significantly reduces prediction errors compared to the individual use of an analytical model or a standalone NN model, particularly when the available measurement data for training is limited. To further enhance the performance of our models, we incorporate NN regularization techniques and employ ensemble averaging, which enable us to achieve an RMSE of less than $1$ dB for a fabricated photonic chip while only using $25\%$ of the available data. The utilization of TL-assisted NNs is promising for addressing the practical limitations associated with data-driven PIC models, particularly concerning experimental data acquisition. \hl{Whereas a scalability study beyond a 3$\times$3 matrix is still required to confirm how the data reduction scales with matrix size, it is expected that the ability to address data scarcity will} be especially crucial when dealing with larger and more intricate MZI mesh architectures for optical matrix multiplication.

\begin{backmatter}
\bmsection{Funding} Villum Foundations, Villum YI, OPTIC-AI (no. VIL29344), ERC CoG FRECOM (no. 771878), National Natural Science Foundation of China (no. 62205114), the Key R\&D Program of Hubei Province (no. 2022BAA001).

\bmsection{Disclosures} The authors declare no conflicts of interest.

\bmsection{Data availability} Data underlying the results presented in this paper are not publicly available at this time but may be obtained from the authors upon reasonable request.


\end{backmatter}

\bibliography{sample}

\bibliographyfullrefs{sample}

\end{document}